\pdfoutput=1
\documentclass[11pt,a4paper]{article}
\usepackage[hyperref]{eacl2021}
\usepackage{times}
\usepackage{latexsym}

\usepackage{microtype}

\aclfinalcopy 
\def\aclpaperid{131} 

\usepackage{amsmath}
\usepackage{amssymb}
\usepackage{arydshln}
\usepackage{booktabs}
\usepackage{tabularx}
\usepackage{tikz}
\usepackage{pgfplots}
\usepackage[normalem]{ulem}
\usepackage{xcolor}
\usepackage{dashbox}%
\usepackage{multirow}
\usepackage{textcomp}
\usepackage{tcolorbox}
\usepackage[ruled, vlined, linesnumbered]{algorithm2e}
\usepackage{mathtools}
\usepackage{soul}

\pgfplotsset{compat=1.13}

\definecolor{plot1}{RGB}{93,147,191}
\definecolor{plot2}{RGB}{233,72,73}
\definecolor{plot3}{RGB}{113,191,110}
\definecolor{plot4}{RGB}{163,73,151}
\definecolor{plot5}{RGB}{230,130,50}
\definecolor{decentgrey}{RGB}{232,232,232}

\usetikzlibrary{calc,fit,positioning,arrows}

\newtcbox{\pattern}{on line,colback=decentgrey,colframe=white,size=fbox,arc=3pt, box align=base, before upper=\strut, top=-2pt, bottom=-2pt, boxrule=0pt}

\newcommand{\bftab}{\fontseries{b}\selectfont}
\newcommand\mask{\_\_\_\_}

\pgfdeclarelayer{bg}    
\pgfsetlayers{bg,main}  

\title{Exploiting Cloze Questions for Few Shot Text Classification and Natural Language Inference}

\author{Timo Schick \\ Sulzer GmbH \\ Munich, Germany \\ \texttt{timo.schick@sulzer.de} \And Hinrich Sch\"utze 
\\ Center for Information and Language Processing \\ LMU Munich, Germany \\ \texttt{inquiries@cislmu.org} \\
}

\author{Timo Schick$^{1,2}$ \quad Hinrich Sch\"utze$^{1}$ \\[0.5em]
	$^{1}$ Center for Information and Language Processing, LMU Munich, Germany \\
	$^{2}$ Sulzer GmbH, Munich, Germany \\[0.5em]
	{\tt schickt@cis.lmu.de}
}

\date{}

\newcounter{notecounter}

\newcommand{\enoteson}{\long\gdef\enote##1##2{{
\stepcounter{notecounter}
{\large\bf
\hspace{1cm}\arabic{notecounter} $<<<$ ##1: ##2
$>>>$\hspace{1cm}}}}}
\enoteson

\begin{document}

\maketitle

\begin{abstract}

Some NLP tasks can be solved in a fully unsupervised fashion
by providing a pretrained language model with ``task descriptions'' in
natural language \citep[e.g.,][]{radford2018language}. While this approach underperforms its
supervised counterpart, we show in this work that the two
ideas can be combined: We introduce Pattern-Exploiting
Training (\textsc{Pet}), a semi-supervised training
procedure that reformulates input examples as cloze-style
phrases to help language models understand a given task. These phrases are
then used to assign soft labels to a large set of unlabeled
examples. Finally, standard supervised training is
performed on the resulting training set. For several tasks and languages,
\textsc{Pet} outperforms supervised
training and strong semi-supervised approaches in low-resource
settings by a large margin.\footnote{Our implementation is publicly available at \url{https://github.com/timoschick/pet}.}

\end{abstract}

\section{Introduction}

Learning from examples is the predominant approach for many NLP tasks: A model is trained on a set of labeled examples from which it then generalizes to unseen data.
Due to the vast number of languages, domains and tasks and the
cost of annotating data, it is common
in real-world uses of NLP
to have only a small number of labeled examples, making \emph{few-shot learning} a highly important research area.
Unfortunately, applying standard supervised learning to small training sets
often performs poorly; 
many problems are difficult to grasp from just looking at a few examples.
For instance, assume we are given the following pieces of text:
\begin{itemize}
\item $T_1$: This was the best pizza I've ever had. 
\item $T_2$: You can get better sushi for half the price.
\item $T_3$: Pizza was average. Not worth the price.
\end{itemize}
Furthermore, imagine we are told that the labels of $T_1$
and $T_2$ are $l$ and $l'$, respectively, and we are
asked to infer the correct label for $T_3$. Based only on
these examples, this is impossible  because plausible justifications can be found for both $l$ and $l'$. However, if we know that the underlying task is to identify whether the text says anything about prices, we can easily assign $l'$ to $T_3$.
This illustrates that solving a task from only a few
examples becomes much easier when we also have
a \emph{task description}, i.e., 
a textual explanation that helps us \emph{understand} what
the task is about.

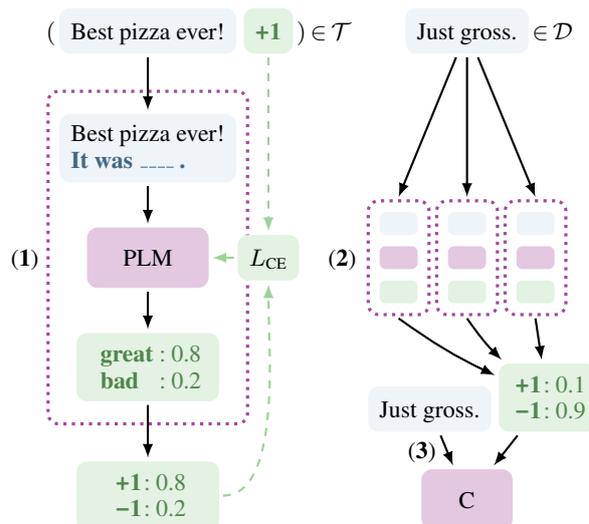
\begin{figure}
\tikzset{
  mnode/.style={fill=plot4!30, rounded corners, minimum width=1.6cm, minimum height=0.8cm, inner sep=0.1cm, outer sep=0.05cm},
  minimodel/.style={fill=plot4!30, rounded corners=0.08cm, minimum width=0.5cm, minimum height=0.2cm, outer sep=0.05cm, inner sep=0},
  minioutput/.style={fill=plot3!20, rounded corners=0.08cm, minimum width=0.5cm, minimum height=0.2cm, outer sep=0.05cm, inner sep=0},
  minipattern/.style={fill=plot1!10, rounded corners=0.08cm, minimum width=0.5cm, minimum height=0.2cm, outer sep=0.05cm, inner sep=0},
  smnode/.style={draw=plot4, dotted, rounded corners, minimum width=0.7cm, minimum height=1.2cm, inner sep=0.1cm, outer sep=0.05cm, very thick},
  textnode/.style={fill=plot1!10, rounded corners, inner sep=0.15cm, outer sep=0.05cm},
  pnode/.style={rounded corners, outer sep=0.05cm, thick,
   text height=1.5ex, text depth=0.25ex, inner ysep=0.3cm, inner xsep=0.2cm, fill=plot3!20, text=plot3!70!black},
  dnode/.style={fill=plot1!10, rounded corners},
  tarrow/.style={draw,->,>=latex, thick},
  tnode/.style={fill=plot1!10, circle, inner sep=0.05cm, outer sep=0.05cm},
  marrow/.style={draw,->,>=latex,dashed},
  explainnode/.style={},
  every node/.style={font=\small, text height=1.5ex, text depth=0.25ex},
}
\centering
\begin{tikzpicture}
\def\tmspacing{0.8cm}
\def\mtspacing{1.6cm}
\def\dspacing{0.2cm}



\node[textnode](input){Best pizza ever!}; 
\node[textnode, right=0cm of input, fill=plot3!20, thick](in-label){\color{plot3!70!black}{\textbf{\texttt{+}1}}};

\node[right=0cm of in-label, outer sep=0, inner sep=0]{$)\,{\in}\,\mathcal{T}$};
\node[left=0cm of input, outer sep=0, inner sep=0]{$($};

\node[textnode, below=0.7cm of input.south west, text height=3.25ex, align=left, anchor=north west](p-of-input){Best pizza ever!\\ \color{plot1!70!black}\textbf{It was \mask{}\ .}};

\node[mnode, below=0.5cm of p-of-input](model){PLM};


\node[pnode, below=0.5cm of model](scores){
\setlength\tabcolsep{1pt}
\begin{tabularx}{1.4cm}{Xcr}
\textbf{great} & : & 0.8 \\
\textbf{bad} & : & 0.2
\end{tabularx}
};
\node[pnode, below=0.7cm of scores, inner xsep=0.4cm](probs){
\setlength\tabcolsep{1pt}
\begin{tabularx}{1cm}{Xcr}
\textbf{\texttt{+}1} & : & 0.8 \\
\textbf{\texttt{-}1} & : & 0.2 \\
\end{tabularx}
};

\node[below=0cm of model.center -| in-label.center, anchor=center, textnode, fill=plot3!20](loss){$L_\text{CE}$};

\begin{pgfonlayer}{bg}    
    \node[draw=plot4, dotted, rounded corners, inner ysep=0.25cm, inner xsep=0.1cm, very thick, fit=(p-of-input)(scores)](model-bb){};
\end{pgfonlayer}
\node[left=0.1cm of model-bb, thick, inner sep=0cm, outer sep=0](step2){ ({\textbf{1}})};

\path[] (input)            edge[tarrow] (input.center |- p-of-input.north);
\path[] (p-of-input)       edge[tarrow] (model);
\path[] (model)            edge[tarrow] (scores);
\path[] (scores)           edge[tarrow] (probs);
\path[] (probs)            edge[tarrow, draw=plot3!70, dashed, out=0, in=270, looseness = 0.7] (loss);
\path[] (in-label)         edge[tarrow, draw=plot3!70, dashed] (loss);
\path[] (loss)             edge[tarrow, draw=plot3!70, dashed] (model);



\node[minimodel, right=4.2cm of model.center, anchor=center](mm1){};
\node[minimodel, left=0.3cm of mm1](mm2){};
\node[minimodel, right=0.3cm of mm1](mm3){};

\node[minipattern, above=0.05cm of mm1](mp1){};
\node[minipattern, above=0.05cm of mm2](mp2){};
\node[minipattern, above=0.05cm of mm3](mp3){};

\node[minioutput, below=0.05cm of mm1](mo1){};
\node[minioutput, below=0.05cm of mm2](mo2){};
\node[minioutput, below=0.05cm of mm3](mo3){};

\node[smnode, fit=(mm1)(mp1)(mo1)](smodel2){};
\node[smnode, fit=(mm2)(mp2)(mo2)](smodel1){};
\node[smnode, fit=(mm3)(mp3)(mo3)](smodel3){};
\node[left=0cm of smodel1, inner sep=0.1cm](){(\textbf{2})};

\node[textnode,above=0cm of smodel2 |- input.center, anchor=center](input2){Just gross.\vphantom{p}};
\node[right=0cm of input2, outer sep=0, inner sep=0]{${\in}\,\mathcal{D}$};

\node[textnode, below=0.85cm of smodel1.south west, anchor=north west](input2-copy){Just gross.\vphantom{p}}; 
\node[pnode, right=0.cm of input2-copy.south east, anchor=south west, minimum width=0, inner xsep=0.1cm](probs2){
\setlength\tabcolsep{1pt}
\begin{tabularx}{1cm}{Xcr}
\textbf{\texttt{+}1} & : & 0.1 \\
\textbf{\texttt{-}1} & : & 0.9 \\
\end{tabularx}
};

\node[mnode, below=0cm of smodel2.center |- probs.south, anchor=south, minimum width=1.2cm](final-classifier){C};


\path[] (input2) edge[tarrow] (smodel1.north);
\path[] (input2) edge[tarrow] (smodel2.north);
\path[] (input2) edge[tarrow] (smodel3.north);

\path[] (smodel1.south) edge[tarrow, bend right=5] (probs2);
\path[] (smodel2.south) edge[tarrow, bend right=5] (probs2);
\path[] (smodel3.south) edge[tarrow] (probs2);

\path[] (input2-copy) edge[tarrow] node[midway, left]{(\textbf{3})} (final-classifier);
\path[] (probs2) edge[tarrow] (final-classifier);

\end{tikzpicture}
\caption{\textsc{Pet} for sentiment
classification. \textbf{(1)} A number of patterns encoding some form of task description are created to
convert training examples to cloze questions;
for each pattern,
a pretrained language model is finetuned. \textbf{(2)} The ensemble of trained models annotates unlabeled data. \textbf{(3)}~A classifier is trained on the resulting soft-labeled dataset.}
\label{pet-idea}
\end{figure}

With the rise of pretrained language models (PLMs) such as GPT
\citep{radford2018improving}, BERT \citep{devlin2018bert} and RoBERTa \citep{liu2019roberta}, the idea of providing
task descriptions has become feasible for neural
architectures: We can simply append such descriptions in
natural language to an input and let the PLM predict
continuations that solve the task
\citep{radford2018language,puri2019zeroshot}. So far, this idea has mostly been considered in zero-shot scenarios where no training data is available at all.

In this work, we show that providing task descriptions can successfully be combined with standard supervised learning in few-shot settings: We introduce \textbf{P}attern-\textbf{E}xploiting \textbf{T}raining (\textsc{Pet}), a semi-supervised training procedure that uses natural language patterns to reformulate input examples into cloze-style phrases. 
As illustrated in Figure~\ref{pet-idea}, \textsc{Pet} works in three steps: First, for each pattern a
separate PLM is finetuned on a small training
set $\mathcal{T}$. The ensemble of all models is then used to annotate a
large unlabeled dataset $\mathcal{D}$ with soft labels. Finally, a standard
classifier is trained on the soft-labeled dataset. We also devise i\textsc{Pet}, an iterative variant of \textsc{Pet} in which this process is repeated with increasing training set sizes.

On a diverse set of tasks in multiple languages, we show
that given a small to medium number of labeled examples, \textsc{Pet} and i\textsc{Pet} substantially outperform unsupervised approaches, supervised training and strong semi-supervised baselines.

\section{Related Work}

\citet{radford2018language} provide hints in the form of natural language patterns for zero-shot learning of challenging tasks such as reading comprehension and question answering (QA). This idea has been applied to unsupervised text classification \citep{puri2019zeroshot}, commonsense knowledge mining \citep{davison-etal-2019-commonsense} and argumentative relation classification \citep{opitz2019argumentative}. \citet{srivastava-etal-2018-zero} use task descriptions for zero-shot classification but require a semantic parser. 
For relation extraction, \citet{bouraoui2020inducing} automatically identify patterns that express given relations. \citet{mccann2018natural} rephrase several tasks as QA problems. \citet{raffel2019exploring} frame various problems as language modeling tasks, but their patterns only loosely resemble natural language and are unsuitable for few-shot learning.\footnote{For example, they convert inputs $(a,b)$ for recognizing textual entailment (RTE) to ``rte sentence1: $a$ sentence2: $b$'', and the PLM is asked to predict strings like ``not\_entailment''.}

Another recent line of work uses cloze-style phrases to probe the
knowledge that PLMs acquire during
pretraining; this includes probing for
factual and commonsense knowledge \citep{trinh2018simple,Petroni_2019,wang-etal-2019-make,sakaguchi2019winogrande}, linguistic capabilities \citep{ettinger2020bert,kassner2019negated},
understanding of rare words \citep{schick2019ota}, and
ability to perform symbolic reasoning
\citep{talmor2019olmpics}. \citet{jiang2019know} consider the problem of finding the
best pattern to express a given task.

Other approaches for few-shot learning in NLP include exploiting examples from related tasks \citep{yu-etal-2018-diverse,gu-etal-2018-meta,dou-etal-2019-investigating,qian-yu-2019-domain,yin2019benchmarking} and using data augmentation \citep{xie2019unsupervised,chen2020mixtext}; the latter commonly relies on back-translation \citep{sennrich-etal-2016-improving}, requiring large amounts of parallel data. Approaches using textual class descriptors typically assume that abundant examples are available for a subset of classes \citep[e.g.,][]{romera2015embarrassingly,veeranna2016using,ye-etal-2020-zero}. In contrast, our approach requires no additional labeled data and provides an intuitive interface to leverage task-specific human knowledge. 

The idea behind i\textsc{Pet} -- training multiple generations of models on data labeled by previous generations -- bears resemblance to self-training and bootstrapping approaches for word sense disambiguation \citep{yarowsky-1995-unsupervised}, relation extraction \citep{brin1999extracting,agichtein2000snowball,batista-etal-2015-semi}, parsing \citep{mcclosky-etal-2006-effective,reichart-rappoport-2007-self,huang-harper-2009-self}, machine translation \citep{hoang-etal-2018-iterative}, and sequence generation \citep{He2020Revisiting}.

\section{Pattern-Exploiting Training}

Let $M$ be a masked language model with vocabulary $V$ and
mask token $\mask{} \in V$, and let $\mathcal{L}$ be a set
of labels for our target classification task $A$. We write
an input for task $A$ as
a sequence of phrases 
$\mathbf{x} = (s_1, \ldots, s_k)$ with $s_i \in V^*$; for example, 
$k=2$ if $A$ is
textual inference (two input sentences).
We define a \emph{pattern} to be a function $P$
that takes $\mathbf{x}$ as input
and outputs a 
phrase or sentence $P(\mathbf{x}) \in V^*$ that contains exactly one
mask token, i.e., its output can be viewed as a cloze
question. Furthermore, we define a \emph{verbalizer} as an
injective function $v: \mathcal{L} \rightarrow V$ that maps
each label to a word from $M$'s vocabulary.
We refer to $(P, v)$ as a \emph{pattern-verbalizer pair} (PVP).

Using a PVP $(P,v)$ enables us to solve task $A$ as follows: Given an input $\mathbf{x}$, we apply $P$ to obtain an input representation $P(\mathbf{x})$, which is then processed by $M$ to determine the label $y \in \mathcal{L}$ for which $v(y)$ is the most likely substitute for the mask. For example, consider the task of identifying whether two sentences $a$ and $b$ contradict each other (label $y_0$) or agree with each other ($y_1$). For this task, we may choose the pattern
$
P(a,b) = \pattern{$a$\text{?} \text{\mask{}, }$b$.}
$
combined with a verbalizer $v$ that maps $y_0$ to ``Yes'' and $y_1$ to ``No''.
Given an example input pair 
\[
\mathbf{x} = (\text{Mia likes pie, Mia hates pie}),
\] the task now changes from having to assign a label without inherent meaning  to answering whether the most likely choice for the masked position in
\[
P(\mathbf{x}) = \pattern{Mia likes pie? \mask{}, Mia hates pie.}
\]
is ``Yes'' or ``No''.

\subsection{PVP Training and Inference}
\label{pvp-training}

Let $\mathbf{p} = (P,v)$ be a PVP. We assume access to a small training set $\mathcal{T}$ and a (typically much larger) set of unlabeled examples $\mathcal{D}$.
For each sequence $\mathbf{z} \in V^*$ that contains exactly one mask token and $w \in V$, we denote with $M(w \mid \mathbf{z})$ the unnormalized score that the language model assigns to $w$ at the masked position. Given some input $\mathbf{x}$, we define the  score for label $l \in \mathcal{L}$ as
\begin{equation*}
s_\mathbf{p}(l \mid \mathbf{x}) = M(v(l) \mid P(\textbf{x}))
\end{equation*}
and obtain a probability distribution over labels using
softmax:
\begin{equation*}
q_\mathbf{p}(l \mid \mathbf{x}) =  \frac{e^{s_\mathbf{p}(l \mid \mathbf{x})}}{ \sum_{l' \in \mathcal{L}} e^{s_\mathbf{p}(l' \mid \mathbf{x})}} \label{eq-softmax}
\end{equation*}
We use the cross-entropy between $q_\mathbf{p}(l \mid
\mathbf{x})$ and the true (one-hot) distribution of training
example $(\mathbf{x}, l)$ -- summed over all $(\mathbf{x},
l) \in \mathcal{T}$ -- as loss for finetuning $M$ for $\mathbf{p}$.

\subsection{Auxiliary Language Modeling}
In our application scenario, only a few training examples are
available and
catastrophic forgetting can occur.
As a PLM finetuned for some PVP is still a language model
at its core, we address this by using language modeling as
auxiliary task. 
With $L_\text{CE}$ denoting cross-entropy loss and $L_\text{MLM}$ language modeling loss, we compute the final loss~as
\begin{equation*}
L = (1 - \alpha) \cdot L_\text{CE} + \alpha \cdot L_\text{MLM}
\end{equation*}
This idea was recently applied by \citet{chronopoulou-etal-2019-embarrassingly} in a data-rich scenario. As $L_\text{MLM}$ is typically much larger than $L_\text{CE}$, in preliminary experiments, we found a small value of $\alpha = 10^{-4}$ to consistently give good results, so we use it in all our experiments. To obtain sentences for language modeling, we use the unlabeled set $\mathcal{D}$. However, we do not train directly on each $\mathbf{x} \in \mathcal{D}$, but rather on $P(\mathbf{x})$, where we never ask the language model to predict anything for the 
masked slot.
\subsection{Combining PVPs}

A key challenge for our approach is that in the absence of a large development set, it is hard to identify which PVPs perform well.
To address this, we use a strategy similar to knowledge distillation \citep{hinton2015distilling}. First, we define a set $\mathcal{P}$ of PVPs that intuitively make sense for a given task $A$. We then use these PVPs as follows:

\begin{enumerate}
\item[\textbf{(1)}] We finetune a separate language model $M_\mathbf{p}$
  for each $\mathbf{p} \in \mathcal{P}$ as described in Section~\ref{pvp-training}. As
  $\mathcal{T}$ is small, this finetuning is cheap even for a large number of PVPs.
\item[\textbf{(2)}]  We use the ensemble $\mathcal{M} = \{ M_\mathbf{p} \mid \mathbf{p} \in \mathcal{P} \}$ of finetuned models to annotate examples from $\mathcal{D}$. We first combine the unnormalized class scores for each example $\mathbf{x} \in \mathcal{D}$ as
\[
{s}_\mathcal{M}(l \mid \mathbf{x}) = \frac{1}{Z} \sum_{\mathbf{p} \in \mathcal{P}} w(\mathbf{p}) \cdot s_\mathbf{p}(l \mid \mathbf{x})
\]
where $Z = \sum_{\mathbf{p} \in \mathcal{P}} w(\mathbf{p})$
and the $w(\mathbf{p})$ are weighting terms for the PVPs.
We experiment with two different realizations of this
weighing term: either we simply set $w(\mathbf{p}) = 1$ for
all $\mathbf{p}$ or we set $w(\mathbf{p})$ to be the
accuracy obtained using $\mathbf{p}$ on the training set
\emph{before} training. We refer to these two variants as
\emph{uniform} and \emph{weighted}.
\citet{jiang2019know} use a similar idea in a zero-shot setting.


We transform the above scores into a probability
  distribution $q$ using softmax. Following
  \citet{hinton2015distilling}, we use a temperature of
  $T=2$ to obtain a suitably soft distribution. All pairs $(\mathbf{x}, q)$ are collected in a (soft-labeled) training set $\mathcal{T}_C$.
  
\item[\textbf{(3)}]  We finetune a PLM $C$ with a standard sequence classification head on $\mathcal{T}_C$.
\end{enumerate}

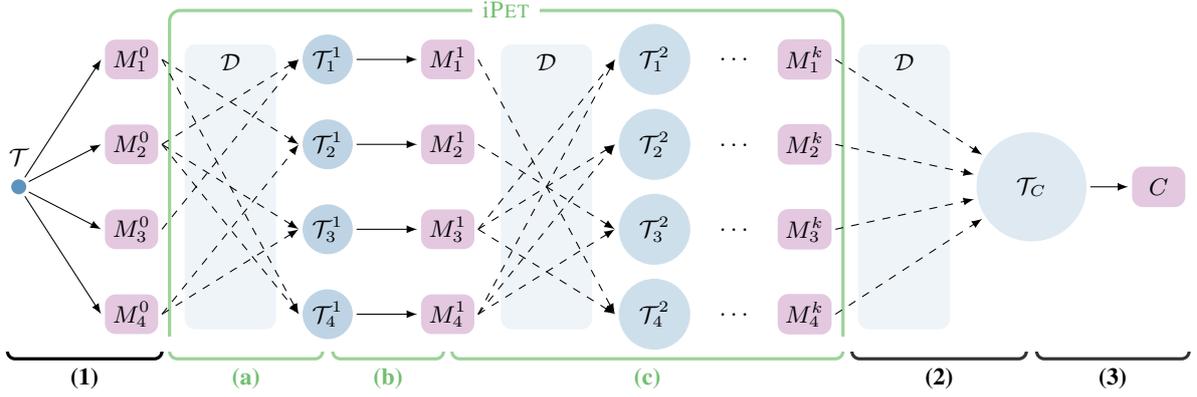
\begin{figure*}
\tikzset{
  mnode/.style={fill=plot4!30, rounded corners, minimum width=0.7cm, inner sep=0.1cm, outer sep=0.05cm},
  tnode/.style={fill=plot1!10, circle, inner sep=0.05cm, outer sep=0.05cm},
  tstart/.style={fill=plot1, circle, inner sep=0.07cm, outer sep=0.05cm},
  dnode/.style={fill=plot1!10, inner xsep=-0.3cm, outer xsep=0.2cm, inner ysep=0.2cm, rounded corners},
  tarrow/.style={draw,->,>=latex},
  marrow/.style={draw,->,>=latex,dashed},
  every node/.style={font=\small},
}
\centering
\begin{tikzpicture}
\def\tmspacing{0.8cm}
\def\mtspacing{1.8cm}
\def\dspacing{0.2cm}

\node[mnode](m1-0){$M_1^0$};
\node[mnode, below=0.5cm of m1-0](m2-0){$M_2^0$};
\node[mnode, below=0.5cm of m2-0](m3-0){$M_3^0$};
\node[mnode, below=0.5cm of m3-0](m4-0){$M_4^0$};
\node[fit=(m1-0)(m4-0)](m-0){};

\node[tstart, left=\tmspacing of m-0](t-init){};
\node[above=0cm of t-init]{$\mathcal{T}$};
\draw [tarrow] (t-init) -- (m1-0.west);
\draw [tarrow] (t-init) -- (m2-0.west);
\draw [tarrow] (t-init) -- (m3-0.west);
\draw [tarrow] (t-init) -- (m4-0.west);

\node[tnode, right=\mtspacing of m1-0,fill=plot1!40](t1-1){$\mathcal{T}_1^1$};
\node[tnode, right=\mtspacing of m2-0,fill=plot1!40](t2-1){$\mathcal{T}_2^1$};
\node[tnode, right=\mtspacing of m3-0,fill=plot1!40](t3-1){$\mathcal{T}_3^1$};
\node[tnode, right=\mtspacing of m4-0,fill=plot1!40](t4-1){$\mathcal{T}_4^1$};

\node[dnode, fit=(m1-0.east)(m4-0.east)(t2-1.west)](d-box0){};
\node[below=0cm of d-box0.north, anchor=north]{$\mathcal{D}$};

\draw [marrow] (m2-0.east) -- (t1-1.west);
\draw [marrow] (m3-0.east) -- (t1-1.west);
\draw [marrow] (m1-0.east) -- (t2-1.west);
\draw [marrow] (m4-0.east) -- (t2-1.west);
\draw [marrow] (m2-0.east) -- (t3-1.west);
\draw [marrow] (m4-0.east) -- (t3-1.west);
\draw [marrow] (m1-0.east) -- (t4-1.west);
\draw [marrow] (m2-0.east) -- (t4-1.west);

\node[mnode, right=\tmspacing of t1-1](m1-1){$M_1^1$};
\node[mnode, right=\tmspacing of t2-1](m2-1){$M_2^1$};
\node[mnode, right=\tmspacing of t3-1](m3-1){$M_3^1$};
\node[mnode, right=\tmspacing of t4-1](m4-1){$M_4^1$};
\draw [tarrow] (t1-1) -- (m1-1);
\draw [tarrow] (t2-1) -- (m2-1);
\draw [tarrow] (t3-1) -- (m3-1);
\draw [tarrow] (t4-1) -- (m4-1);

\node[tnode, right=\mtspacing of m1-1, inner sep=0.15cm, fill=plot1!30](t1-2){$\mathcal{T}_1^2$};
\node[tnode, right=\mtspacing of m2-1, inner sep=0.15cm, fill=plot1!30](t2-2){$\mathcal{T}_2^2$};
\node[tnode, right=\mtspacing of m3-1, inner sep=0.15cm, fill=plot1!30](t3-2){$\mathcal{T}_3^2$};
\node[tnode, right=\mtspacing of m4-1, inner sep=0.15cm, fill=plot1!30](t4-2){$\mathcal{T}_4^2$};

\node[dnode, fit=(m1-1.east)(m4-1.east)(t2-2.west)](d-box1){};
\node[below=0cm of d-box1.north, anchor=north]{$\mathcal{D}$};

\draw [marrow] (m3-1.east) -- (t1-2.west);
\draw [marrow] (m4-1.east) -- (t1-2.west);
\draw [marrow] (m3-1.east) -- (t2-2.west);
\draw [marrow] (m4-1.east) -- (t2-2.west);
\draw [marrow] (m2-1.east) -- (t3-2.west);
\draw [marrow] (m4-1.east) -- (t3-2.west);
\draw [marrow] (m1-1.east) -- (t4-2.west);
\draw [marrow] (m3-1.east) -- (t4-2.west);

\node[right=\dspacing of t1-2](d1){$\ldots$};
\node[right=\dspacing of t2-2](d2){$\ldots$};
\node[right=\dspacing of t3-2](d3){$\ldots$};
\node[right=\dspacing of t4-2](d4){$\ldots$};

\node[mnode, right=\dspacing of d1](m1-k){$M_1^k$};
\node[mnode, right=\dspacing of d2](m2-k){$M_2^k$};
\node[mnode, right=\dspacing of d3](m3-k){$M_3^k$};
\node[mnode, right=\dspacing of d4](m4-k){$M_4^k$};
\node[fit=(m1-k)(m4-k), inner sep=0](m-k){};

\node[tnode, right=\mtspacing of m-k, inner sep=0.35cm, fill=plot1!20](t-final){$\mathcal{T}_C$};
\node[mnode, right=0.5cm of t-final](m-final){$C\vphantom{_2^k}$};

\node[dnode, fit=(m1-k.east)(m4-k.east)(t-final.west)](d-box2){};
\node[below=0cm of d-box2.north, anchor=north]{$\mathcal{D}$};

\begin{pgfonlayer}{bg}    
\node[fit=(d-box0.west)(d-box2.west)(t1-2)(t4-2), draw=plot3!70, very thick, rounded corners, inner xsep=0, outer xsep=0](ipet-box){};
\node[fill=white, inner sep=0.25cm, outer sep=0, yshift=-0.45cm, fit=(d-box0.south west)(d-box2.south west)]{};
\node[above=0cm of ipet-box, anchor=center, fill=white, text=plot3]{{i\textsc{Pet}}};
\end{pgfonlayer}

\draw [marrow] (m1-k.east) -- (t-final);
\draw [marrow] (m2-k.east) -- (t-final);
\draw [marrow] (m3-k.east) -- (t-final);
\draw [marrow] (m4-k.east) -- (t-final);
\draw [tarrow] (t-final) -- (m-final);

\begin{pgfonlayer}{bg}    
\node[draw=black, inner xsep=0, outer sep=0, yshift=-0.15cm, very thick, rounded corners=.1cm, fit=(t-init.west |- m4-0.south) (m4-0.south east)](brace1){};
\node[below=0.1cm of brace1, text=black, inner sep=0, outer sep=0]{\strut\textbf{(1)}};

\node[draw=plot3!70, inner xsep=0, outer sep=0, yshift=-0.15cm, very thick, rounded corners=.1cm, fit=(ipet-box.west |- m4-0.south) (t4-1.260 |- m4-0.south)](bracea){};
\node[below=0.1cm of bracea, text=plot3, inner sep=0, outer sep=0]{\strut\textbf{(a)}};

\node[draw=plot3!70, inner xsep=0, outer sep=0, yshift=-0.15cm, very thick, rounded corners=.1cm, fit=(t4-1.280 |- m4-1.south)(m4-1.260)](braceb){};
\node[below=0.1cm of braceb, text=plot3, inner sep=0, outer sep=0]{\strut\textbf{(b)}};

\node[draw=plot3!70, inner xsep=0, outer sep=0, yshift=-0.15cm, very thick, rounded corners=.1cm, fit=(m4-1.280 |- m4-1.south)(ipet-box.east |- m4-1.south)](bracec){};
\node[below=0.1cm of bracec, text=plot3, inner sep=0, outer sep=0]{\strut\textbf{(c)}};

\node[inner sep=0, outer sep=0, right=0.1cm of ipet-box.east](brace2dummy){};
\node[draw=black, opacity=0.8, inner xsep=0, outer sep=0, yshift=-0.15cm, very thick, rounded corners=.1cm, fit=(brace2dummy |- m4-1.south)(t-final.265 |- m4-1.south)](brace2){};
\node[below=0.1cm of brace2,text=black, inner sep=0, outer sep=0]{\strut\textbf{(2)}};

\node[draw=black, opacity=0.8, inner xsep=0, outer sep=0, yshift=-0.15cm, very thick, rounded corners=.1cm, fit=(t-final.275 |- m4-1.south)(m-final.east |- m4-1.south)](brace3){};
\node[below=0.1cm of brace3,text=black, inner sep=0, outer sep=0]{\strut\textbf{(3)}};

\node[fill=white, fit=(brace1.west)(brace3.east)(brace1.north), inner xsep=0.05cm, inner ysep=0.03cm, outer sep=0]{};
\end{pgfonlayer}

\end{tikzpicture}
\caption{Schematic representation of \textsc{Pet} (1-3) and i\textsc{Pet} (a-c). \textbf{(1)} The initial training set is used to finetune an ensemble of PLMs. \textbf{(a)} For each model, a random subset of other models generates a new training set by labeling examples from $\mathcal{D}$. \textbf{(b)} A new set of \textsc{Pet} models is trained using the larger, model-specific datasets. \textbf{(c)} The previous two steps are repeated $k$ times, each time increasing the size of the generated training sets by a factor of $d$. \textbf{(2)} The final set of models is used to create a soft-labeled dataset $\mathcal{T}_C$. \textbf{(3)} A classifier $C$ is trained on this dataset.}
\label{ipet-schema}
\end{figure*}

The finetuned model $C$ then serves as our classifier for $A$. All steps described above are depicted in Figure~\ref{ipet-schema}; an example is shown in Figure~\ref{pet-idea}.

\subsection{Iterative \textsc{Pet} (i\textsc{Pet})}
\label{ipet}

Distilling the knowledge of all
individual models 
into a single classifier $C$ means
they
cannot learn from each
other. As some patterns  perform (possibly much)
worse than others,  the training set
$\mathcal{T}_C$ for our final model may therefore contain many
mislabeled examples.

To compensate for this shortcoming, we devise i\textsc{Pet}, an iterative variant of \textsc{Pet}. The core idea of i\textsc{Pet} is to train several \emph{generations} of models on datasets of increasing size. To this end, we first enlarge the original dataset $\mathcal{T}$ by labeling selected examples from $\mathcal{D}$ using a random subset of trained \textsc{Pet} models (Figure~\ref{ipet-schema}a). We then train a new generation of \textsc{Pet} models on the enlarged dataset~(b); this process is repeated several times (c).

More formally, let $\mathcal{M}^0 = \{ M_1^0, \ldots, M_n^0 \}$ be the initial set of \textsc{Pet} models finetuned on $\mathcal{T}$, where each $M_i^0$ is trained for some PVP $\mathbf{p}_i$. We train $k$ generations of models $\mathcal{M}^1, \ldots, \mathcal{M}^k$ where $\mathcal{M}^j = \{ M_1^j, \ldots, M_n^j \}$ and each $M_i^j$ is trained for $\mathbf{p}_i$ on its own training set $\mathcal{T}_i^j$. In each iteration, we multiply the training set size by a fixed constant $d \in \mathbb{N}$ while maintaining the label ratio of the original dataset. That is, with $c_0(l)$ denoting the number of examples with label $l$ in $\mathcal{T}$, each $\mathcal{T}_i^j$ contains $c_j(l) = d \cdot c_{j-1}(l)$ examples with label $l$. 
This is achieved by generating each $\mathcal{T}_i^j$ as follows:

\begin{enumerate}
\item We obtain $\mathcal{N} \subset \mathcal{M}^{j-1} \setminus \{ M_i^{j-1} \}$ by randomly choosing $\lambda \cdot (n-1)$ models from the previous generation with $\lambda \in (0,1]$ being a hyperparameter.
\item Using this subset, we create a labeled dataset 
\[
{\mathcal{T}}_\mathcal{N} = \{ (\mathbf{x}, \arg\max_{l \in \mathcal{L}} s_\mathcal{N} (l \mid \mathbf{x})) \mid \mathbf{x} \in \mathcal{D} \}\,.\]
For each $l \in \mathcal{L}$, we obtain $\mathcal{T}_\mathcal{N}(l) \subset \mathcal{T}_\mathcal{N}$ by randomly choosing $c_j(l) - c_0(l)$ examples with label $l$ from $\mathcal{T}_\mathcal{N}$. To avoid training future generations on mislabeled data, we prefer examples for which the ensemble of models is confident in its prediction. The underlying intuition is that even without calibration, examples for which labels are predicted with high confidence are typically more likely to be classified correctly \citep{guo2017calibration}. Therefore, when drawing from $\mathcal{T}_\mathcal{N}$, we set the probability of each $(\mathbf{x}, y)$ proportional to $s_\mathcal{N}(l \mid \mathbf{x})$.
\item We define $\mathcal{T}_i^j
= \mathcal{T} \cup \bigcup_{l \in \mathcal{L}} \mathcal{T}_\mathcal{N}(l)$.
As can easily be verified, this dataset contains $c_j(l)$ examples for each $l \in \mathcal{L}$.
\end{enumerate}
After training $k$ generations of \textsc{Pet} models, we use $\mathcal{M}^k$ to create $\mathcal{T}_C$ and train $C$ as in basic \textsc{Pet}.

With minor adjustments, i\textsc{Pet} can even be used in a
zero-shot setting. To this end, we define $\mathcal{M}^0$ to
be the set of \emph{untrained} models and
$c_1(l) = {10 / |\mathcal{L}|}$
for all $l \in \mathcal{L}$ so that $\mathcal{M}^1$ is trained on 10 examples evenly distributed across all labels. As $\mathcal{T}_\mathcal{N}$ may not contain enough examples for some label $l$, we create all $\mathcal{T}_\mathcal{N}(l)$ by sampling from the 100 examples $\mathbf{x} \in \mathcal{D}$ for which $s_\mathcal{N}(l \mid x)$ is the highest, even if $l \neq \arg\max_{l \in \mathcal{L}} s_\mathcal{N}(l \mid x)$.
For each subsequent generation, we proceed exactly as in basic i\textsc{Pet}.

\section{Experiments}

We evaluate \textsc{Pet} on four English datasets: Yelp Reviews, AG's News, Yahoo Questions \citep{zhang2015character} and MNLI \citep{williams2018mnli}. 
Additionally, we use x-stance \citep{vamvas2020xstance} to investigate how well \textsc{Pet} works for other languages. 
For all experiments
on English,
we use RoBERTa large \citep{liu2019roberta} as language model; for x-stance, we use XLM-R \citep{conneau2019unsupervised}. 
We investigate the performance of \textsc{Pet} and all
baselines for different training set sizes;
each model is trained three times using different seeds and average results are reported.

As we consider a few-shot setting, we assume no access to a
large development set on which hyperparameters could be
optimized. Our choice of hyperparameters is thus based on
choices made in previous work and practical
considerations. 
We use a learning rate of $1\cdot10^{-5}$, a
batch size of $16$ and a maximum sequence length of $256$.
Unless otherwise specified, we always use the weighted variant of \textsc{Pet} with auxiliary language modeling.
For i\textsc{Pet}, we set $\lambda = 0.25$ and $d = 5$; that is, we select $25\%$ of all models to label examples for the next generation and quintuple the number of training examples in each iteration. We train new generations until each model was trained on at least $1000$ examples, i.e., we set $k = \lceil \log_d(1000 / |\mathcal{T}|) \rceil$.
As we always repeat training three times, the ensemble $\mathcal{M}$ (or $\mathcal{M}^0$) for $n$ PVPs contains $3n$ models.
Further hyperparameters and detailed explanations for all our choices are given in Appendix~B.

\subsection{Patterns}
\label{patterns}

We now describe the patterns and verbalizers used for all
tasks. We use two vertical bars ($\|$) to mark boundaries
between  text segments.\footnote{The way different segments
  are handled depends on the model being used; they
  may e.g. be assigned different embeddings
  \citep{devlin2018bert} or separated by special tokens
  \citep{liu2019roberta,yang2019xlnet}.
  For example, 
  ``{$a$ $\|$ $b$}'' is given to
  BERT as the input
``[CLS] $a$ [SEP] $b$ [SEP]''.}

\paragraph{Yelp}

For the Yelp Reviews Full Star dataset \citep{zhang2015character}, the task is to estimate the rating that a customer gave to a restaurant on a $1$- to $5$-star scale based on their review's text. We define the following patterns for an input text $a$:
\begin{align*}
P_1(a) & = \pattern{It was \mask{}. $a$} & P_2(a) & = \pattern{Just \mask{}! $\|$ $a$} \\
P_3(a) & = \rlap{\pattern{$a$. All in all, it was \mask{}.}}\\
P_4(a) & = \rlap{\pattern{$a$ $\|$ In summary, the restaurant is \mask{}.}}
\end{align*}
We define a single verbalizer $v$ for all patterns as
\begin{align*}
v(1) & = \text{terrible} & v(2) & = \text{bad} & v(3) & = \text{okay} \\
v(4) & = \text{good} & v(5) & = \text{great}\,
\end{align*}

\paragraph{AG's News}

AG's News is a news classification dataset, where given a headline $a$ and text body $b$, news have to be classified as belonging to one of the categories \emph{World} ($1$), \emph{Sports} ($2$), \emph{Business} ($3$) or \emph{Science/Tech} ($4$). For $\mathbf{x} = (a,b)$, we define the following patterns:
\begin{align*}
P_1(\mathbf{x}) & = \pattern{\mask{}: $a$ $b$} & P_2(\mathbf{x}) & = \pattern{$a$ ( \mask{} ) $b$} \\
P_3(\mathbf{x}) & = \pattern{\mask{} -- $a$ $b$} & P_4(\mathbf{x}) & = \pattern{$a$ $b$ ( \mask{} )} \\
P_5(\mathbf{x}) & = \rlap{\pattern{\mask{} News: $a$ $b$}} \\
P_6(\mathbf{x}) & = \rlap{\pattern{[ Category: \mask{} ] $a$ $b$}}
\end{align*}
We use a verbalizer that maps $1$--$4$ to ``World'', ``Sports'', ``Business'' and ``Tech'', respectively.

\paragraph{Yahoo}

Yahoo Questions \citep{zhang2015character} is a text
classification dataset. Given a question $a$ and an answer
$b$, one of ten possible categories has to be assigned. We
use the same patterns as for AG's News, but we replace the
word ``News'' in $P_5$ with the word ``Question''.
We define a verbalizer that maps
categories $1$--$10$ to ``Society'', ``Science'', ``Health'',
``Education'', ``Computer'', ``Sports'', ``Business'',
``Entertainment'', ``Relationship'' and ``Politics''.

\paragraph{MNLI}

The MNLI dataset \citep{williams2018mnli} consists of text
pairs $\mathbf{x} = (a,b)$.
The task is to find out whether $a$ implies $b$  ($0$), $a$
and $b$ contradict each other ($1$) or neither ($2$). We define
\begin{align*}
P_1(\mathbf{x})\,{=}\,\pattern{``$a$''? $\|$ \mask{}, ``$b$''} && P_2(\mathbf{x})\,{=}\,\pattern{$a$? $\|$ \mask{}, $b$}
\end{align*}
and consider two different verbalizers $v_1$ and $v_2$:
\begin{alignat*}{3}
v_1(0) & = \text{Wrong}\ \ & v_1(1) & = \text{Right}\ \ & v_1(2) & = \text{Maybe}\\
v_2(0) & = \text{No} & v_2(1) & = \text{Yes} & v_2(2) & = \text{Maybe}
\end{alignat*}
Combining the two patterns with the two verbalizers results in a total of 4 PVPs.

\begin{table*}
\small
\centering
\newcolumntype{Y}{>{\centering\arraybackslash}X}
\begin{tabularx}{0.87\linewidth}{cllYYYc}
\toprule
\textbf{Line}&\textbf{Examples} & \textbf{Method} & \multicolumn{1}{c}{\textbf{Yelp}} & \multicolumn{1}{c}{\textbf{AG's}} & \multicolumn{1}{c}{\textbf{Yahoo}} & {\textbf{MNLI (m/mm)}} \\
\midrule
\phantom{1}1&\multirow{3}{*}{$|\mathcal{T}| = 0$} 
&
unsupervised
(avg)  & 33.8 \scriptsize $\pm$9.6          & 69.5 \scriptsize $\pm$7.2          & 44.0 \scriptsize $\pm$9.1          & 39.1 {\scriptsize $\pm$4.3} / 39.8 \scriptsize $\pm$5.1 \\
\phantom{1}2&&
unsupervised
(max)  & 40.8 \scriptsize $\pm$0.0          & 79.4 \scriptsize $\pm$0.0          & 56.4 \scriptsize $\pm$0.0          & 43.8 {\scriptsize $\pm$0.0} / 45.0 \scriptsize $\pm$0.0 \\
\phantom{1}3&& i\textsc{Pet} & {\bftab 56.7} \scriptsize $\pm$0.2 & {\bftab 87.5} \scriptsize $\pm$0.1 & {\bftab 70.7} \scriptsize $\pm$0.1 & {\bftab 53.6} {\scriptsize $\pm$0.1} / {\bftab 54.2} \scriptsize $\pm$0.1 \\
\midrule

\phantom{1}4&\multirow{3}{*}{$|\mathcal{T}| = 10$}
& supervised    & 21.1 \scriptsize $\pm$1.6          & 25.0 \scriptsize $\pm$0.1          & 10.1 \scriptsize $\pm$0.1          & 34.2 {\scriptsize $\pm$2.1} / 34.1 \scriptsize $\pm$2.0 \\
\phantom{1}5&& \textsc{Pet}  & 52.9 \scriptsize $\pm$0.1          & 87.5 \scriptsize $\pm$0.0          & 63.8 \scriptsize $\pm$0.2          & 41.8 {\scriptsize $\pm$0.1} / 41.5 \scriptsize $\pm$0.2 \\
\phantom{1}6&& i\textsc{Pet} & {\bftab 57.6} \scriptsize $\pm$0.0 & {\bftab 89.3} \scriptsize $\pm$0.1 & {\bftab 70.7} \scriptsize $\pm$0.1 & {\bftab 43.2} {\scriptsize $\pm$0.0} / {\bftab 45.7} \scriptsize $\pm$0.1 \\
\midrule

\phantom{1}7&\multirow{3}{*}{$|\mathcal{T}| = 50$}
& supervised    & 44.8 \scriptsize $\pm$2.7          & 82.1 \scriptsize $\pm$2.5          & 52.5 \scriptsize $\pm$3.1          & 45.6 {\scriptsize $\pm$1.8} / 47.6 \scriptsize $\pm$2.4 \\
\phantom{1}8&& \textsc{Pet}  & 60.0 \scriptsize $\pm$0.1          & 86.3 \scriptsize $\pm$0.0          & 66.2 \scriptsize $\pm$0.1          & 63.9 {\scriptsize $\pm$0.0} / 64.2 \scriptsize $\pm$0.0 \\
\phantom{1}9&& i\textsc{Pet} & {\bftab 60.7} \scriptsize $\pm$0.1 & {\bftab 88.4} \scriptsize $\pm$0.1 & {\bftab 69.7} \scriptsize $\pm$0.0 & {\bftab 67.4} {\scriptsize $\pm$0.3} / {\bftab 68.3} \scriptsize $\pm$0.3 \\
\midrule

10&\multirow{3}{*}{$|\mathcal{T}| = 100$}
& supervised    & 53.0 \scriptsize $\pm$3.1          & 86.0 \scriptsize $\pm$0.7          & 62.9 \scriptsize $\pm$0.9          & 47.9 {\scriptsize $\pm$2.8} / 51.2 \scriptsize $\pm$2.6 \\
11&& \textsc{Pet}  & 61.9 \scriptsize $\pm$0.0          & 88.3 \scriptsize $\pm$0.1          & 69.2 \scriptsize $\pm$0.0          & 74.7 {\scriptsize $\pm$0.3} / 75.9 \scriptsize $\pm$0.4 \\
12&& i\textsc{Pet} & {\bftab 62.9} \scriptsize $\pm$0.0 & {\bftab 89.6} \scriptsize $\pm$0.1 & {\bftab 71.2} \scriptsize $\pm$0.1 & {\bftab 78.4} {\scriptsize $\pm$0.7} / {\bftab 78.6} \scriptsize $\pm$0.5 \\
\midrule

13&\multirow{2}{*}{$|\mathcal{T}| = 1000$}
& supervised    & 63.0 \scriptsize $\pm$0.5          & {\bftab 86.9} \scriptsize $\pm$0.4 & 70.5 \scriptsize $\pm$0.3          & 73.1 {\scriptsize $\pm$0.2} / 74.8 \scriptsize $\pm$0.3 \\
14&& \textsc{Pet}  & {\bftab 64.8} \scriptsize $\pm$0.1 & {\bftab 86.9} \scriptsize $\pm$0.2 & {\bftab 72.7} \scriptsize $\pm$0.0 & {\bftab 85.3} {\scriptsize $\pm$0.2} / {\bftab 85.5} \scriptsize $\pm$0.4 \\
\bottomrule
\end{tabularx}
\caption{Average accuracy and standard deviation for RoBERTa
(large) on Yelp, AG's News, Yahoo and MNLI
(m:matched/mm:mismatched) for five training set sizes
$|\mathcal{T}|$.}
\label{main_results}
\end{table*}

\paragraph{X-Stance}

The x-stance dataset \citep{vamvas2020xstance} is a multilingual stance detection dataset with German, French and Italian examples. Each example $\mathbf{x} = (a,b)$ consists of a question $a$ concerning some political issue and a comment $b$; the task is to identify whether the writer of $b$ supports the subject of the question ($0$) or not ($1$). We use two simple patterns
\begin{align*}
P_1(\mathbf{x}) = \pattern{``$a$'' $\|$ \mask{}. ``$b$''} && P_2(\mathbf{x}) = \pattern{$a$ $\|$ \mask{}. $b$}
\end{align*}
and define an English verbalizer $v_\text{En}$ mapping $0$
to ``Yes'' and $1$ to ``No'' as well as a French (German)
verbalizer $v_\text{Fr}$ ($v_\text{De}$), replacing
``Yes''
and ``No'' with
 ``Oui'' and
``Non'' (``Ja'' and ``Nein''). We do not define an Italian verbalizer because
x-stance does not contain any Italian training examples.
 
\subsection{Results}

\paragraph{English Datasets} Table~\ref{main_results} shows results for English text classification and language understanding tasks; we report mean
accuracy and standard deviation for three training runs.
Lines 1--2 (L1--L2) show unsupervised performance, i.e., individual PVPs without \emph{any} training \citep[similar to][]{radford2018improving,puri2019zeroshot};
we give both average results across all PVPs (avg)
and results for the PVP that works best on the test set (max). 
The large difference between both rows highlights the importance of coping with the fact that without looking at the test set, we have no means of evaluating which PVPs perform well.
Zero-shot i\textsc{Pet}
clearly outperforms the unsupervised baselines for all datasets (L3
vs L1); on AG's News, it even performs better than
standard supervised training with 1000 examples (L3 vs L13).
With just 10 training examples, standard supervised learning
does not perform above chance (L4). In
contrast, \textsc{Pet} (L5) performs much better than the fully
unsupervised baselines (L1--L2); training multiple generations using
i\textsc{Pet} (L6) gives consistent improvements. 
As we increase the training set size, the
performance gains of \textsc{Pet} and i\textsc{Pet} become smaller, but for
both 50 and 100 examples, \textsc{Pet} continues to
considerably outperform standard supervised training
(L8 vs L7, L11 vs L10)
with i\textsc{Pet} (L9, L12) still giving consistent
improvements.
For $|\mathcal{T}| = 1000$,
\textsc{Pet} 
has no advantage on AG's  but still improves accuracy
for all other tasks (L14 vs L13)%
.\footnote{One of the three supervised
	MNLI runs for $|\mathcal{T}|=1000$ underfitted the training data and performed
	extremely poorly. This run is excluded in the reported score
	(73.1/74.8).}

\begin{table}
	\small
	\centering
	\newcolumntype{Y}{>{\centering\arraybackslash}X}
	\begin{tabularx}{\linewidth}{llYYYc}
		\toprule
		\textbf{Ex.} & \textbf{Method} & \multicolumn{1}{c}{\textbf{Yelp}} & \multicolumn{1}{c}{\textbf{AG's}} & \multicolumn{1}{c}{\textbf{Yahoo}} & \multicolumn{1}{c}{\textbf{MNLI}}  \\
		\midrule
		\multirow{4}{*}{\rotatebox[origin=c]{90}{$|\mathcal{T}| = 10$}} 
		& UDA           & 27.3 & 72.6 & 36.7 & 34.7  \\
		& MixText       & 20.4 & 81.1 & 20.6 & 32.9 \\
		& \textsc{Pet}  & 48.8 & 84.1 & 59.0 & 39.5 \\
		& i\textsc{Pet} & \bftab 52.9 & \bftab 87.5 & \bftab 67.0 & \bftab 42.1 \\
		\midrule
		\multirow{4}{*}{\rotatebox[origin=c]{90}{$|\mathcal{T}| = 50$}} 
		& UDA           & 46.6 & 83.0 & 60.2 & 40.8 \\
	    & MixText       & 31.3 & 84.8 & 61.5 & 34.8 \\
	    & \textsc{Pet}  & 55.3 & 86.4 & 63.3 & 55.1 \\
		& i\textsc{Pet} & \bftab 56.7 & \bftab 87.3 & \bftab 66.4 & \bftab 56.3 \\		
		\bottomrule
	\end{tabularx}
	\caption{Comparison of \textsc{Pet} with two state-of-the-art semi-supervised methods using RoBERTa (base)}
	\label{sota-results}
\end{table}

\paragraph{Comparison with SotA}

We compare \textsc{Pet} to UDA \citep{xie2019unsupervised} and MixText \citep{chen2020mixtext}, two state-of-the-art methods for semi-supervised learning in NLP that rely on data augmentation. Whereas \textsc{Pet} requires that a task can be expressed using patterns and that such patterns be found, UDA and MixText both use backtranslation \citep{sennrich-etal-2016-improving} and thus require thousands of labeled examples for training a machine translation model.
We use RoBERTa (base) for our comparison as MixText is specifically tailored towards a 12-layer Transformer \citep{Vaswani2017}. Both \citet{xie2019unsupervised} and \citet{chen2020mixtext} use large development sets to optimize the number of training steps. We instead try several values for both approaches directly on the test set and only report the \emph{best} results obtained. Despite this, Table~\ref{sota-results} shows that \textsc{Pet} and i\textsc{Pet} substantially outperform both methods across all tasks, clearly demonstrating the benefit of incorporating human knowledge in the form of PVPs. 
 
\paragraph{X-Stance} We evaluate \textsc{Pet} on x-stance to
investigate (i) whether it works for
languages other than English and (ii) whether it also
brings improvements
when training sets have medium size.
In contrast to  \citet{vamvas2020xstance}, we do not perform any hyperparameter optimization on  dev and use a shorter maximum sequence length (256 vs 512) to speed up training and evaluation. 

\begin{table}
	\small
	\centering
	\newcolumntype{Y}{>{\centering\arraybackslash}X}
	\begin{tabularx}{\linewidth}{llYYY}
		\toprule
		\textbf{Examples} & \textbf{Method} & \multicolumn{1}{c}{\textbf{De}} & \multicolumn{1}{c}{\textbf{Fr}} & \multicolumn{1}{c}{\textbf{It}} \\
		\midrule
		
		\multirow{2}{*}{$|\mathcal{T}| = 1000$}
		& supervised   & $43.3$          & $49.5$          & $41.0$  \\
		& \textsc{Pet} & $\mathbf{66.4}$ & $\mathbf{68.7}$ & $\mathbf{64.7}$  \\
		\midrule
		
		\multirow{2}{*}{$|\mathcal{T}| = 2000$}
		& supervised   & $57.4$          & $62.1$          & $52.8$  \\
		& \textsc{Pet} & $\mathbf{69.5}$ & $\mathbf{71.7}$ & $\mathbf{67.3}$  \\
		\midrule
		
		\multirow{2}{*}{$|\mathcal{T}| = 4000$}
		& supervised   & $63.2$          & $66.7$          & $58.7$ \\
		& \textsc{Pet} & $\mathbf{71.7}$ & $\mathbf{74.0}$ & $\mathbf{69.5}$  \\
		\midrule
		
		\multirow{2}{*}{$\mathcal{T}_\text{De}$ , $\mathcal{T}_\text{Fr}$}
		& supervised   & $76.6$          & $76.0$          & $71.0$ \\
		& \textsc{Pet} & $\mathbf{77.9}$ & $\mathbf{79.0}$ & $\mathbf{73.6}$ \\
		\midrule
		
		\multirow{3}{*}{$\mathcal{T}_\text{De} + \mathcal{T}_\text{Fr}$}
		& sup. (*)        & $76.8$          & $76.7$          & $70.2$ \\
		& supervised   & $77.6$          & $79.1$          & $75.9$ \\
		& \textsc{Pet} & $\mathbf{78.8}$ & $\mathbf{80.6}$ & $\mathbf{77.2}$ \\
		
		\bottomrule
	\end{tabularx}
	\caption{Results on x-stance intra-target for XLM-R (base) trained on subsets of $\mathcal{T}_\text{De}$ and $\mathcal{T}_\text{Fr}$ and for joint training on all data ($\mathcal{T}_\text{De} + \mathcal{T}_\text{Fr}$). (*): Best results for mBERT reported in \citet{vamvas2020xstance}.}
	\label{xstance-results}
\end{table}

To investigate whether \textsc{Pet} brings benefits even
when numerous examples are available, we consider training
set sizes of $1000$, $2000$, and $4000$; for each of these
configurations, we separately finetune French and German 
models to allow for a more straightforward downsampling of
the training data. Additionally, we train models on the
entire
French ($|\mathcal{T}_\text{Fr}| = 11\,790$)
and
German ($|\mathcal{T}_\text{De}| = 33\,850$)
training
sets. In this case we do not have any additional unlabeled
data, so we simply set $\mathcal{D} = \mathcal{T}$. For the
French
models, we use
$v_\text{En}$ and $v_\text{Fr}$
as
verbalizers and for
German
$v_\text{En}$ and $v_\text{De}$
(Section \ref{patterns}). Finally, we also investigate the
performance of a model trained jointly on French and German data ($|\mathcal{T}_\text{Fr} + \mathcal{T}_\text{De}| = 45\,640$) using $v_\text{En}$,  $v_\text{Fr}$ and $v_\text{De}$.

Results are shown in Table~\ref{xstance-results};
following \citet{vamvas2020xstance}, we report the
macro-average of the F1 scores for labels~0 and~1, averaged over three runs.
For Italian (column~``It''), we report the
average zero-shot cross-lingual performance of German and
French models as there are no Italian training examples. Our
results show that \textsc{Pet} brings huge improvements
across all languages even when training on much more than a
thousand examples; it also considerably improves zero-shot
cross-lingual performance.

\section{Analysis}
\label{analysis}

\begin{table}
	\small
	\newcolumntype{Y}{>{\centering\arraybackslash}X}
	\begin{tabularx}{\linewidth}{lYYYc}
		\toprule
		\textbf{Method} & \textbf{Yelp} & \textbf{AG's} & \textbf{Yahoo} & \textbf{MNLI} \\
		\midrule
		min & 39.6 & 82.1 & 50.2 & 36.4  \\
		max & 52.4 & 85.0  & 63.6 & 40.2 \\
		\textsc{Pet} (no distillation) & 51.7 & 87.0 & 62.8 & 40.6 \\
		\textsc{Pet} uniform & 52.7 & 87.3 & \textbf{63.8} & \textbf{42.0} \\
		\textsc{Pet} weighted & \textbf{52.9} & \textbf{87.5} & \textbf{63.8} & 41.8 \\
		\bottomrule
	\end{tabularx}
	
	\caption{Minimum (min) and maximum (max) accuracy of models
		based on individual PVPs as well as \textsc{Pet} with and without knowledge distillation
		($|\mathcal{T}| = 10$).}
	\label{pattern-scores}
\end{table}

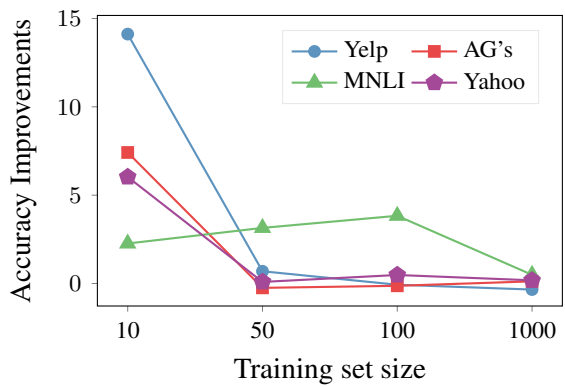
\begin{figure}
\centering
\begin{tikzpicture}
\begin{axis}[
	cycle list name=color list,
	xlabel={Training set size},
		ylabel={Accuracy Improvements},
    ymin = -0.2,
    ymax = 14.1,
    xmin = 1,
    xmax = 4,
    enlarge x limits={0.075},
    enlarge y limits={0.075},
    xtick = {1, 2, 3, 4},
    xticklabels = {10, 50, 100, 1000},
    xtick pos=left,
    ytick pos=left,
    ylabel near ticks,
    xlabel near ticks,
    tick align=outside,
    major tick length=0.075cm,
    width = \linewidth,
    height = 0.22\textheight,
    x tick label style={/pgf/number format/1000 sep=},
    legend style={draw=decentgrey, at={(0.95,0.95)},anchor=north east, font=\small},
    legend cell align=left,
    legend columns=2,
    tick label style={font=\footnotesize}
]

\addplot[mark=*, thick, mark options={solid}, plot1] coordinates {
(1,14.116)
(2,0.693)
(3,-0.068)
(4,-0.342)
};
\addlegendentry{Yelp}

\addplot[mark=square*, thick, mark options={solid}, plot2] coordinates {
(1,7.417)
(2,-0.2455)
(3,-0.129)
(4,0.1228)
};
\addlegendentry{AG's}

\addplot[mark=triangle*, mark size=3pt, thick, mark options={solid}, plot3] coordinates {
(1,2.265)
(2,3.148)
(3,3.834)
(4,0.493)
};
\addlegendentry{MNLI}

\addplot[mark=pentagon*, mark size=3pt, thick, mark options={solid}, plot4] coordinates {
(1,6.03)
(2,0.09)
(3,0.483)
(4,0.181)
};
\addlegendentry{Yahoo}


\end{axis}
\end{tikzpicture}
\caption{Accuracy improvements for \textsc{Pet} due to adding $L_\text{MLM}$ during training}
\label{language-modeling}
\end{figure}

\paragraph{Combining PVPs}
We first investigate whether \textsc{Pet} is able to cope with
situations were some PVPs perform much worse
than others.
For $|\mathcal{T}| = 10$, Table~\ref{pattern-scores}
compares the performance of \textsc{Pet} to that of the 
best and worst performing patterns after finetuning; we also include results obtained using the ensemble of \textsc{Pet} models corresponding to individual PVPs without knowledge distillation.
Even after
finetuning, the gap between the best and worst pattern is
large, especially for Yelp.
However, \textsc{Pet} is not only able to compensate
for this, but even improves accuracies over using only the
best-performing pattern across all tasks. Distillation brings consistent improvements over the ensemble; additionally, it significantly reduces the size of the final classifier. We find no
clear difference between the uniform and weighted
variants of \textsc{Pet}.

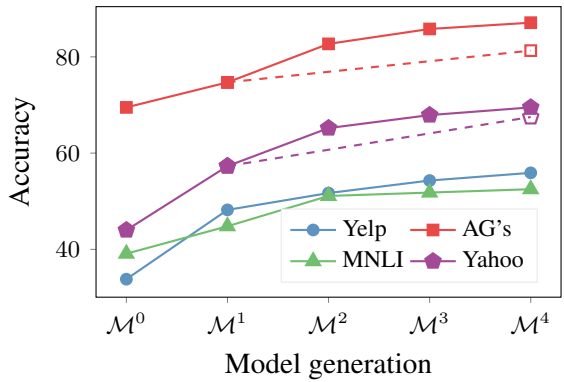
\begin{figure}
\centering
\begin{tikzpicture}
\begin{axis}[
    table/col sep=comma,
	cycle list name=color list,
	xlabel={Model generation},
		ylabel={Accuracy},
    ymin = 34,
    ymax = 86.5,
    xmin = 0,
    xmax = 4,
    enlarge x limits={0.075},
    enlarge y limits={0.075},
    xtick = {0,1,2,3,4},
    xticklabels = {$\mathcal{M}^0$,$\mathcal{M}^1$,$\mathcal{M}^2$,$\mathcal{M}^3$,$\mathcal{M}^4$},
    xtick pos=left,
    ytick pos=left,
    ylabel near ticks,
    xlabel near ticks,
    tick align=outside,
    major tick length=0.075cm,
    width = \linewidth,
    height = 0.22\textheight,
    x tick label style={/pgf/number format/1000 sep=},
    legend style={draw=decentgrey, at={(0.95,0.05)},anchor=south east, font=\small},
    legend cell align=left,
    legend columns=2,
    tick label style={font=\footnotesize}
]

\addplot[mark=*, thick, mark options={solid}, plot1] 
 table [x=generation, y=yelp, col sep=comma] {zero-shot-scores.csv};
\addlegendentry{Yelp}

\addplot[mark=square*, thick, mark options={solid}, plot2] 
 table [x=generation, y=ags, col sep=comma] {zero-shot-scores.csv};
\addlegendentry{AG's}

\addplot[mark=triangle*, mark size=3pt, thick, mark options={solid}, plot3] 
 table [x=generation, y=mnli, col sep=comma] {zero-shot-scores.csv};
\addlegendentry{MNLI}

\addplot[mark=pentagon*, mark size=3pt, thick, mark options={solid}, plot4] 
 table [x=generation, y=yahoo, col sep=comma] {zero-shot-scores.csv};
\addlegendentry{Yahoo}

\addplot[mark=square, thick, mark options={solid}, dashed, plot2] coordinates {
(1,74.7)
(4,81.3)
};

\addplot[mark=pentagon, thick, mark size=3pt, mark options={solid}, dashed, plot4] coordinates {
(1,57.3)
(4,67.5)
};

\end{axis}
\end{tikzpicture}
\caption{Average accuracy for each generation of models with i\textsc{Pet} in a zero-shot setting. Accuracy on AG's News and Yahoo when skipping generation 2 and 3 is indicated through dashed lines.}
\label{ipet-analysis}
\end{figure}

\paragraph{Auxiliary Language Modeling} We analyze the
influence of the auxiliary language modeling task
on \textsc{Pet}'s
performance. Figure~\ref{language-modeling} shows
performance improvements from adding the language modeling
task for four training set sizes. We see that
the auxiliary task is extremely valuable when training on just 10 examples. With more data, it becomes less important, sometimes even leading to worse performance. Only for MNLI, we find language modeling to consistently help.

\paragraph{Iterative \textsc{Pet}}

To check whether i\textsc{Pet} is  able to improve
models over multiple generations, Figure~\ref{ipet-analysis}
shows the average performance of all generations of models
in a zero-shot setting.
Each additional iteration does indeed further improve the ensemble's performance. We did not investigate whether continuing this process for even more iterations gives further improvements.

Another natural question is whether similar results can
be obtained with fewer iterations by increasing the training
set size more aggressively. To answer this question, we skip
generations 2 and 3 for AG's News and Yahoo and for both tasks directly let
ensemble $\mathcal{M}^1$ annotate $10\cdot5^4$ examples for
$\mathcal{M}^4$. As indicated in Figure~\ref{ipet-analysis}
through dashed lines, this clearly leads to worse
performance, highlighting the importance of only gradually
increasing the training set size. We surmise that this is
the case because annotating too many examples too early
leads to a large percentage of mislabeled training examples.

\paragraph{In-Domain Pretraining} Unlike our supervised
baseline, \textsc{Pet} makes use of the additional unlabeled
dataset $\mathcal{D}$. Thus, at least some of \textsc{Pet}'s
performance gains over the supervised baseline may 
arise from this additional in-domain data. 

\begin{figure}
\centering
\begin{tikzpicture}
\begin{axis}[
	cycle list name=color list,
	xlabel={Training set size},
		ylabel={Accuracy},
    ymin = 20.2,
    ymax = 66.4,
    xmin = 1,
    xmax = 4,
    enlarge x limits={0.075},
    enlarge y limits={0.075},
    xtick = {1, 2, 3, 4},
    xticklabels = {10, 50, 100, 1000},
    xtick pos=left,
    ytick pos=left,
    ylabel near ticks,
    xlabel near ticks,
    tick align=outside,
    major tick length=0.075cm,
    width = \linewidth,
    height = 0.22\textheight,
    x tick label style={/pgf/number format/1000 sep=},
    legend style={draw=decentgrey, at={(0.95,0.05)},anchor=south east, font=\small},
    legend cell align=left,
    legend columns=1,
    tick label style={font=\footnotesize}
]

\addplot[mark=*, thick, mark options={solid}, plot1] coordinates {
(1,52.9)
(2,60)
(3,61.9)
(4,64.8)
};
\addlegendentry{\textsc{Pet}}

\addplot[mark=o, dashed, thick, mark options={solid}, plot1] coordinates {
(1,58)
(2,62.1)
(3,63.5)
(4,66.4)
};
\addlegendentry{\textsc{Pet} + {PT}}

\addplot[mark=triangle*, mark size=3pt, thick, mark options={solid}, plot2] coordinates {
(1,21.1)
(2,44.8)
(3,53)
(4,63)
};
\addlegendentry{sup.}

\addplot[mark=triangle, dashed, mark size=3pt, thick, mark options={solid}, plot2] coordinates {
(1,20.2)
(2,48)
(3,56.6)
(4,64.8)
};
\addlegendentry{sup. + {PT}}
\end{axis}
\end{tikzpicture}
\caption{Accuracy of supervised learning (sup.) and \textsc{Pet} both with and without pretraining ({PT}) on Yelp}
\label{language-model-pretraining}
\end{figure}
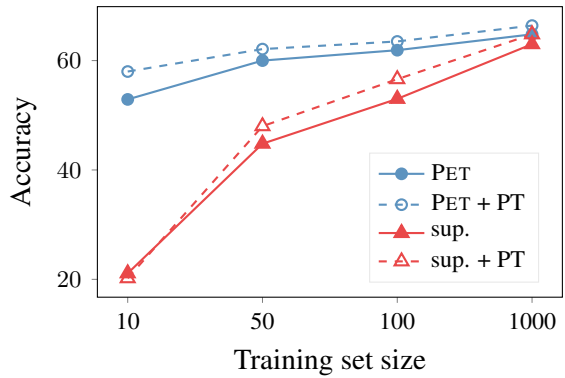

To test this
hypothesis, we simply further pretrain RoBERTa on in-domain
data, a common technique for improving text classification
accuracy \citep[e.g.,][]{howard2018universal,chi2019finetune}. As
language model pretraining is expensive in terms of GPU
usage, we do so only for the Yelp dataset.
Figure~\ref{language-model-pretraining} shows results of
supervised learning and \textsc{Pet} both with and without
this in-domain pretraining. While pretraining does indeed improve accuracy for supervised training,
the supervised model still clearly performs worse
than \textsc{Pet}, showing that the success of our method is
not simply due to the usage of additional unlabeled
data. Interestingly, in-domain pretraining is also helpful
for \textsc{Pet}, indicating that \textsc{Pet} leverages
unlabeled data in a way that is clearly different from  standard masked
language model pretraining.

\section{Conclusion}

We have shown that providing task descriptions to pretrained
language models can be combined with standard supervised
training. Our proposed method, \textsc{Pet}, consists of
defining pairs of cloze question patterns and verbalizers
that help leverage the knowledge contained within pretrained
language models for downstream tasks. We finetune models for
all pattern-verbalizer pairs and use them to create large
annotated datasets on which standard classifiers can be
trained. When the initial amount of training data is
limited, \textsc{Pet} gives large improvements over standard supervised training and strong semi-supervised approaches.

\section*{Acknowledgments}
This work was funded by the European Research Council (ERC \#740516).
We would like to thank the anonymous reviewers
for their helpful comments.

\bibliographystyle{acl_natbib}
\bibliography{literatur}

\clearpage

\appendix

\section{Implementation}

Our implementation of \textsc{Pet} and i\textsc{Pet} is based on the Transformers library \citep{wolf2019transformers} and PyTorch \citep{paszke2017automatic}.

\section{Training Details}
\label{appendix-hyperparameter-choices}

Except for the in-domain pretraining experiment described in Section~5%
, all of our experiments were conducted using a single GPU with 11GB RAM (NVIDIA GeForce GTX 1080 Ti). 

\subsection{Hyperparameter Choices}

Relevant training hyperparameters for both individual \textsc{Pet} models and the final classifier $C$ as well as our supervised baseline are listed in Table~\ref{hyperparameters-table}. All hyperparameters were selected based on the following considerations and experiments:

\paragraph{Batch size / maximum length}
Both batch size and maximum sequence length (or block size) are chosen so that one batch fits into 11GB of GPU memory. As \citet{devlin2018bert} and \citet{liu2019roberta} use larger batch sizes of 16--32, we accumulate gradients for 4 steps to obtain an effective batch size of 16.

\paragraph{Learning rate} We found a learning rate of $5\mathrm{e}{-5}$ (as used by \citet{devlin2018bert}) to often result in unstable training for regular supervised learning with no accuracy improvements on the training set. We therefore use a lower learning rate of $1\mathrm{e}{-5}$, similar to \citet{liu2019roberta}. Experiments with various learning rates can be found in Appendix~\ref{appendix-hyperparameters}.

\paragraph{Training steps} As the number of training epochs recommended by \citet{liu2019roberta} in a data-rich scenario is in the range 2--10, we perform supervised training for 250 training steps, corresponding to 4 epochs when training on 1000 examples. 
For individual \textsc{Pet} models, we
subdivide each batch into one labeled example from
$\mathcal{T}$ to compute $L_\text{CE}$ and three unlabeled
examples from $\mathcal{D}$ to compute
$L_\text{MLM}$. Accordingly, we multiply the number of total
training steps by $4$
(i.e., 1000),
so that the number of times each labeled example is seen
remains constant
($16\cdot 250 = 4 \cdot 1000$). For the final \textsc{Pet} classifier, we train for 5000 steps due to the increased training set size (depending on the task, the unlabeled set $\mathcal{D}$ contains at least $20\,000$ examples). Deviating from the above, we always perform training for 3 epochs on x-stance to match the setup of \citet{vamvas2020xstance} more closely. The effect of varying the number of training steps is further investigated in Appendix~\ref{appendix-hyperparameters}. 

\paragraph{Temperature} We choose a temperature of 2 when training the final classifier following \citet{hinton2015distilling}.

\paragraph{Auxiliary language modeling} To find a suitable value of $\alpha$ for combining language modeling loss and cross-entropy loss, we first observed that in the early stages of training, the former is a few orders of magnitude higher than the latter for all tasks considered. We thus selected a range $\{ 1\mathrm{e}{-}3, 1\mathrm{e}{-}4, 1\mathrm{e}{-}5 \}$ of reasonable choices for $\alpha$ and performed preliminary experiments on Yelp with 100 training examples to find the best value among these candidates. To this end, we split the training examples into a training set and a dev set using both a 90/10 split and a 50/50 split and took the value of $\alpha$ that maximizes average dev set accuracy. We adopt this value for all other tasks and training set sizes without further optimization. 

\paragraph{Models per ensemble} As we always train three models per pattern, for both i\textsc{Pet} and training the final classifier $C$, the ensemble $\mathcal{M}$ (or $\mathcal{M}^0$) for $n$ PVPs contains $3n$ models. This ensures consistency as randomly choosing any of the three models for each PVP would result in high variance. In preliminary experiments, we found this to have only little impact on the final model's performance.

\paragraph{i\textsc{Pet} dataset size}  For i\textsc{Pet}, we quintuple the number of training examples after each iteration ($d = 5$) so that only a small number of generations is required to reach a sufficient amount of labeled data. We did not choose a higher value because we presume that this may cause training sets for early generations to contain a prohibitively large amount of mislabeled data.

\paragraph{i\textsc{Pet} dataset creation} We create training sets for the next generation in i\textsc{Pet} using 25\% of the models in the current generation ($\lambda = 0.25$) because we want the training sets for all models to be diverse while at the same time, a single model should not have too much influence.

\paragraph{Others} For all other hyperparameters listed in Table~\ref{hyperparameters-table}, we took the default settings of the Transformers library \citep{wolf2019transformers}.

\subsection{Number of parameters} 

As \textsc{Pet} does not require any additional learnable parameters, the number of parameters for both \textsc{Pet} and i\textsc{Pet} is identical to the number of parameters in the underlying language model: 355M for RoBERTa (large) and 270M for XLM-R (base).

\subsection{Average runtime} 

Training a single \textsc{Pet} classifier for 250 steps on one GPU took approximately 30 minutes; training for 1000 steps with auxiliary language modeling took $~$60 minutes. Depending on the task, labeling examples from $\mathcal{D}$ took 15--30 minutes per model. Training the final classifier $C$ for 5000 steps on the soft-labeled dataset $\mathcal{T}_C$ took 2 hours on average.

\subsection{Comparison with SotA}

For comparing \textsc{Pet} to UDA \citep{xie2019unsupervised} and MixText \citep{chen2020mixtext}, we reduce the number of unlabeled examples by half to speed up the required backtranslation step. We use the backtranslation script provided by \citet{chen2020mixtext} with their recommended hyperparameter values and use both Russian and German as intermediate languages. 

For MixText, we use the original implementation\footnote{\url{https://github.com/GT-SALT/MixText}} and the default set of hyperparameters. Specifically, each batch consists of 4 labeled and 8 unlabeled examples, we use layers 7, 9 and 12 for mixing, we set $T=5$, $\alpha=16$, and use a learning rate of $5 \cdot 10^{-6}$ for RoBERTa and $5 \cdot 10 ^{-4}$ for the final classification layer. We optimize the number of training steps for each task and dataset size in the range $\{1000, 2000, 3000, 4000, 5000\}$.

For UDA, we use a PyTorch-based reimplementation\footnote{\url{https://github.com/SanghunYun/UDA_pytorch}}. We use the same batch size as for MixText and the hyperparameter values recommended by \citet{xie2019unsupervised}; we use an exponential schedule for training signal annealing and a learning rate of $2\cdot10^{-5}$. We optimize the number of training steps for each task and dataset size in the range $\{500, 1000, 1500, \ldots, 10000\}$.

\subsection{In-Domain Pretraining} 

For in-domain pretraining experiments described in Section~5%
, we use the language model finetuning script of the Transformers library \citep{wolf2019transformers}; all hyperparameters are listed in the last column of Table~\ref{hyperparameters-table}. Pretraining was performed on a total of 3 NVIDIA GeForce GTX 1080 Ti GPUs.

\begin{table*}
	\begin{tabularx}{\linewidth}{lXXXXX}
		\toprule
		\textbf{Parameter} 						& \textsc{Pet} $-$LM & \textsc{Pet} (En/Xs) & $C$ (En/Xs) & sup. (En/Xs) & In-Dom. PT \\
		\midrule
		\texttt{adam\_epsilon} 					& 1e-8 & 1e-8 & 1e-8 & 1e-8 & 1e-8 \\
		* \texttt{alpha} 						& -- & 1e-4 & -- & -- & -- \\
		\texttt{block\_size} 					& -- 	& -- & -- & -- & 256\\
		\texttt{gradient\_accumulation\_steps} 	& 4 & 4 & 4 & 4 & 2 \\
		\texttt{learning\_rate} 				& 1e-5 & 1e-5 & 1e-5 & 1e-5 & 5e-5 \\
		\texttt{max\_grad\_norm} 				& 1.0 & 1.0 & 1.0 & 1.0 & 1.0 \\
		\texttt{max\_seq\_length} 				& 256 & 256 & 256 & 256 & -- \\
		\texttt{max\_steps} 					& 250 	& 1000 / -- & 5000 / -- & 250 / -- & 50000 \\
		\texttt{mlm\_probability} 				& -- & 0.15 & -- & -- & 0.15 \\
		\texttt{num\_train\_epochs}				& -- & -- / 3 & -- / 3 & -- / 3 & -- \\
		\texttt{per\_gpu\_train\_batch\_size}   & 4 & 1 & 4 & 4 & 2 \\
		* \texttt{per\_gpu\_helper\_batch\_size} 	& -- & 3 & -- & -- & -- \\
		* \texttt{temperature} 					& -- & -- & 2.0 & -- & -- \\
		\texttt{weight\_decay} 					& 0.01 & 0.01 & 0.01 & 0.01 & 0.0 \\
		\bottomrule
	\end{tabularx}
	\caption{Hyperparameters for training individual \textsc{Pet} models without auxiliary language modeling (\textsc{Pet}$-$LM) and with language modeling (\textsc{Pet}), the final \textsc{Pet} classifier ($C$), regular supervised training (sup.) and in-domain pretraining (In-Dom. PT). Whenever different values are used for the English datasets (En) and x-stance (Xs), both values are given separated by a slash. (*): \textsc{Pet}-specific hyperparameters}
	\label{hyperparameters-table}
\end{table*}

\section{Dataset Details}

For each task and number of examples $t$, we create the training set $\mathcal{T}$ by collecting the first $t/|\mathcal{L}|$ examples per label from the original training set, where $|\mathcal{L}|$ is the number of labels for the task. Similarly, we construct the set $\mathcal{D}$ of unlabeled examples by selecting $10\,000$ examples per label and removing all labels. For evaluation, we use the official test set for all tasks except MNLI, for which we report results on the dev set; this is due to the limit of 2 submissions per 14 hours for the official MNLI test set. An overview of the number of test examples and links to downloadable versions of all used datasets can be found in Table~\ref{datasets-table}.

\paragraph{Preprocessing} In some of the datasets used, newlines are indicated through the character sequence ``{\textbackslash n}''. As the vocabularies of RoBERTa and XLM-R do not feature a newline, we replace this sequence with a single space. We do not perform any other preprocessing, except shortening all examples to the maximum sequence length of 256 tokens. This is done using the \emph{longest first} strategy implemented in the Transformers library. For \textsc{Pet}, all input sequences are truncated \emph{before} applying patterns.

\paragraph{Evaluation metrics} For Yelp, AG's News, Yahoo and MNLI, we use accuracy. For x-stance, we report macro-average of F1 scores using the evaluation script of \citet{vamvas2020xstance}. 

\begin{table*}
	\begin{tabularx}{\linewidth}{lXr}
		\toprule
		\textbf{Dataset}	    & \textbf{Link}  								& \textbf{Test Examples} \\
		\midrule
		AG's News 			    & \url{http://goo.gl/JyCnZq} 					& 7600 \\
		MNLI (m / mm) 			& \url{https://cims.nyu.edu/~sbowman/multinli/} & 10000 / 10000 \\
		X-Stance (De / Fr / It) & \url{https://github.com/ZurichNLP/xstance} 	& 3479 / 1284 / 1173 \\
		Yahoo! Answers 		    & \url{http://goo.gl/JyCnZq} 					& 60000 \\
		Yelp Review Full 	    & \url{http://goo.gl/JyCnZq} 					& 50000 \\
		\bottomrule
	\end{tabularx}
	\caption{Download links and number of test examples for all datasets}
	\label{datasets-table}
\end{table*}

\section{Hyperparameter Importance}
\label{appendix-hyperparameters}

To analyze the importance of hyperparameter choices for \textsc{Pet}'s performance gains over supervised learning, we look at the influence of both the learning rate ($\mathrm{LR}$) and the number of training steps on their test set accuracies. 

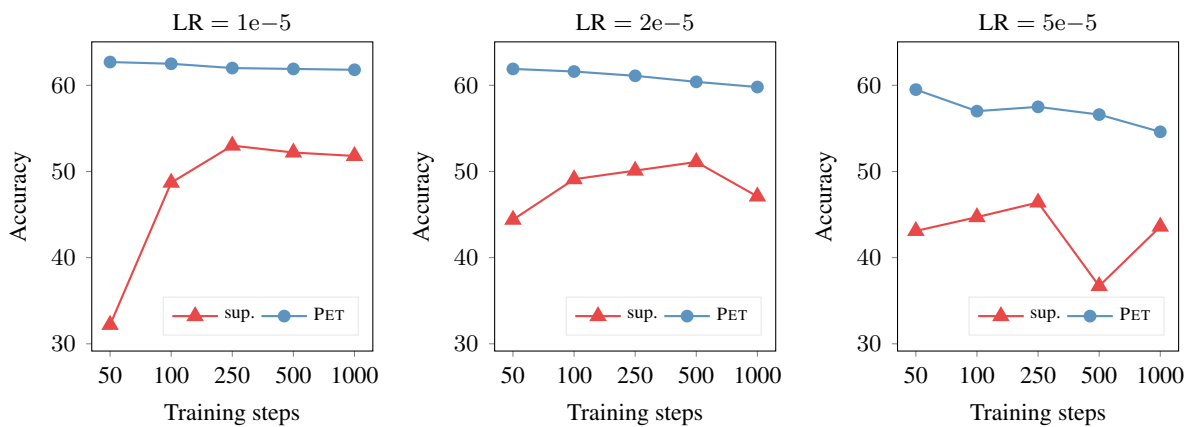
\begin{figure*}
	\centering
	\begin{tikzpicture}
	\begin{axis}[
	title={\small $\text{LR} = 1\mathrm{e}{-5}$},
	title style={yshift=-1.25ex},
	cycle list name=color list,
	xlabel={\small Training steps},
	ylabel={\small Accuracy},
	ymin = 31.5,
	ymax = 62.7,
	xmin = 1,
	xmax = 5,
	enlarge x limits={0.075},
	enlarge y limits={0.075},
	xtick = {1, 2, 3, 4, 5},
	xticklabels = {50, 100, 250, 500, 1000},
	xtick pos=left,
	ytick pos=left,
	ylabel near ticks,
	xlabel near ticks,
	tick align=outside,
	major tick length=0.075cm,
	width = 0.33\linewidth,
	height = 0.23\textheight,
	x tick label style={/pgf/number format/1000 sep=},
	legend style={draw=decentgrey, at={(0.95,0.05)},anchor=south east, font=\scriptsize},
	legend cell align=left,
	legend columns=2,
	tick label style={font=\footnotesize}
	]
	
	\addplot[mark=triangle*,  mark size=3pt, thick, mark options={solid}, plot2] coordinates {
		(1,32.2)
		(2,48.7)
		(3,53.0)
		(4,52.2)
		(5,51.8)
	};
	\addlegendentry{sup.}
	
	\addplot[mark=*, thick, mark options={solid}, plot1] coordinates {
		(1,62.7)
		(2,62.5)
		(3,62.0)
		(4,61.9)
		(5,61.8)
	};
	\addlegendentry{\textsc{Pet}}
	
	\end{axis}
	\end{tikzpicture}
	~%
	\begin{tikzpicture}
	\begin{axis}[
	title={\small $\text{LR} = 2\mathrm{e}{-5}$},
	title style={yshift=-1.25ex},
	cycle list name=color list,
	xlabel={\small Training steps},
	ylabel={\small Accuracy},
	ymin = 31.5,
	ymax = 62.7,
	xmin = 1,
	xmax = 5,
	enlarge x limits={0.075},
	enlarge y limits={0.075},
	xtick = {1, 2, 3, 4, 5},
	xticklabels = {50, 100, 250, 500, 1000},
	xtick pos=left,
	ytick pos=left,
	ylabel near ticks,
	xlabel near ticks,
	tick align=outside,
	major tick length=0.075cm,
	width = 0.33\linewidth,
	height = 0.23\textheight,
	x tick label style={/pgf/number format/1000 sep=},
	legend style={draw=decentgrey, at={(0.95,0.05)},anchor=south east, font=\scriptsize},
	legend cell align=left,
	legend columns=2,
	tick label style={font=\footnotesize}
	]
	
	\addplot[mark=triangle*,  mark size=3pt, thick, mark options={solid}, plot2] coordinates {
		(1,44.4)
		(2,49.1)
		(3,50.1)
		(4,51.1)
		(5,47.1)
	};
	\addlegendentry{sup.}
	
	\addplot[mark=*, thick, mark options={solid}, plot1] coordinates {
		(1,61.9)
		(2,61.6)
		(3,61.1)
		(4,60.4)
		(5,59.8)
	};
	\addlegendentry{\textsc{Pet}}
	
	\end{axis}
	\end{tikzpicture}
	~%
	\begin{tikzpicture}
	\begin{axis}[
	title={\small $\text{LR} = 5\mathrm{e}{-5}$},
	title style={yshift=-1.25ex},
	cycle list name=color list,
	xlabel={\small Training steps},
	ylabel={\small Accuracy},
	ymin = 31.5,
	ymax = 62.7,
	xmin = 1,
	xmax = 5,
	enlarge x limits={0.075},
	enlarge y limits={0.075},
	xtick = {1, 2, 3, 4, 5},
	xticklabels = {50, 100, 250, 500, 1000},
	xtick pos=left,
	ytick pos=left,
	ylabel near ticks,
	xlabel near ticks,
	tick align=outside,
	major tick length=0.075cm,
	width = 0.33\linewidth,
	height = 0.23\textheight,
	x tick label style={/pgf/number format/1000 sep=},
	legend style={draw=decentgrey, at={(0.95,0.05)},anchor=south east, font=\scriptsize},
	legend cell align=left,
	legend columns=2,
	tick label style={font=\footnotesize}
	]
	
	\addplot[mark=triangle*, mark size=3pt, thick, mark options={solid}, plot2] coordinates {
		(1,43.1)
		(2,44.7)
		(3,46.4)
		(4,36.7)
		(5,43.6)
	};
	\addlegendentry{sup.}
	
	\addplot[mark=*, thick, mark options={solid}, plot1] coordinates {
		(1,59.5)
		(2,57.0)
		(3,57.5)
		(4,56.6)
		(5,54.6)
	};
	\addlegendentry{\textsc{Pet}}
	
	\end{axis}
	\end{tikzpicture}
	\caption{Performance of supervised learning and \textsc{Pet} (weighted, without auxiliary language modeling) for various learning rates and training steps on Yelp with 100 training examples}
	\label{hyperparameters-yelp}
\end{figure*}

We try values of $\{ 1\mathrm{e}{-5}, 2\mathrm{e}{-5}, 5\mathrm{e}{-5} \}$ for the learning rate and $\{50, 100, 250, 500, 1000\}$ for the number of training steps. As this results in 30 different configurations for just one task and training set size, we only perform this analysis on Yelp with 100 examples, for which results can be seen in Figure~\ref{hyperparameters-yelp}. For supervised learning, the configuration used throughout the paper ($\mathrm{LR}=1\mathrm{e}{-5}$, 250 steps) turns out to perform best whereas for \textsc{Pet}, training for fewer steps consistently performs even better. Importantly, \textsc{Pet} clearly outperforms regular supervised training regardless of the chosen learning rate and number of training steps.

\section{Automatic Verbalizer Search}
\label{avs}

Given a set of patterns $P_1, \ldots, P_n$, manually finding a verbalization $v(l)$ for each $l \in \mathcal{L}$ that represents the meaning of $l$ well and corresponds to a single token in $V$ can be difficult. We therefore devise \emph{automatic verbalizer search} (AVS), a procedure that automatically finds suitable verbalizers given a training set $\mathcal{T}$ and a language model $M$.

Assuming we already have a PVP $\mathbf{p} = (P, v)$, we can easily check whether some token $t \in V$ is a good verbalization of $l \in \mathcal{L}$. To this end, we define $\mathbf{p}[l\gets t] = (P, v')$, where $v'$ is identical to $v$, except that $v'(l) = t$. Intuitively, if $t$ represents $l$ well, then  $q_{\mathbf{p}[l \gets t]}(l \mid \mathbf{x})$ (i.e., the probability $M$ assigns to $t$ given $P(\mathbf{x})$) should be high only for those examples $(\mathbf{x}, y) \in \mathcal{T}$ where $y = l$. We thus define the score of $t$ for $l$ given $\mathbf{p}$ as
\begin{multline*}
s_{l}(t \mid \mathbf{p}) =  \frac{1}{|\mathcal{T}_l|} \cdot \sum_{{(\mathbf{x}, y) \in \mathcal{T}_l}} q_{\mathbf{p}[l \gets t]}(l \mid \mathbf{x})
\\ - \frac{1}{|\mathcal{T} \setminus \mathcal{T}_{l}|} \cdot \sum_{{(\mathbf{x}, y) \in \mathcal{T} \setminus \mathcal{T}_{l}}} q_{\mathbf{p}[l \gets t]}(l \mid \mathbf{x})
\end{multline*}
where  $\mathcal{T}_l = \{ (\mathbf{x}, y) \in \mathcal{T} : y = l \}$ is the set of all training examples with label $l$.  While this allows us to easily compute the best verbalization for $l$ as \[
\hat{t} = \arg\max_{t \in V} s_l(t \mid \mathbf{p})\,,
\] it requires us to already know verbalizations $v(l')$ for all other labels $l'$.

AVS solves this problem as follows: We first assign random verbalizations to all labels and then repeatedly recompute the best verbalization for each label. As we do not want the resulting verbalizer to depend strongly on the initial random assignment, we simply consider multiple such assignments. Specifically, we define an initial probability distribution $\rho_0$ where for all $t \in V, l \in \mathcal{L}$, $\rho_0(t \mid l) = 1 / |V|$ is the probability of choosing $t$ as verbalization for $l$. For each $l \in \mathcal{L}$, we then sample $k$ verbalizers $v_1, \ldots, v_k$ using $\rho_0$ to compute \[
s_l^k(t) = \frac{1}{n\cdot k} \sum_{i=1}^{n}\sum_{j=1}^k s_l(t \mid (P_i, v_j))
\]
for all $t \in V$.\footnote{Note that the score $s_l^k(t)$ jointly considers all patterns; in preliminary experiments, we found this to result in more robust verbalizers.} These scores enable us to define a probability distribution $\rho_1$ that more closely reflects a word's suitability as a verbalizer for a given label:
\[
\rho_1(t \mid l) = \frac{1}{Z} \max(s_l^k(t), \epsilon)
\]
where $Z = \sum_{t' \in V} \max(s_l^k(t'), \epsilon)$ and $\epsilon \geq 0$ ensures that $\rho_1$ is a proper probability distribution. We repeat this process to obtain a sequence of probability distributions $\rho_1, \ldots, \rho_{i_\text{max}}$.
Finally, we choose the $m \in \mathbb{N}$ most likely tokens according to $\rho_{i_\text{max}}(t \mid l)$ as verbalizers for each $l$. During training and inference, we compute the unnormalized score $s_\mathbf{p}(y \mid \mathbf{x})$ for each label by averaging over its $m$ verbalizers.

We analyze the performance of AVS for all tasks with $|\mathcal{T}| = 50$ training examples and set $k = 250$, $\epsilon = 10^{-3}$, $i_\text{max} = 5$ and $m = 10$.\footnote{We tried values of $k$ and $i_\text{max}$ in $\{250, 500, 1000\}$ and $\{5, 10, 20\}$, respectively, but found the resulting verbalizers to be almost identical.} To speed up the search, we additionally restrict our search space to tokens $t \in V$ that contain at least two alphabetic characters. Of these tokens, we only keep the $10\,000$ most frequent ones in $\mathcal{D}$.

Results are shown in Table~\ref{automatic-verbalizers}. As can be seen, carefully handcrafted verbalizers perform much better than AVS; however, \textsc{Pet} with AVS still considerably outperforms regular supervised training while eliminating the challenge of manually finding suitable verbalizers. Table~\ref{avs-yelp} shows the most probable verbalizers found using AVS for the Yelp dataset. While most verbalizers for this dataset intuitively make sense, we found AVS to struggle with finding good verbalizers for three out of ten labels in the Yahoo dataset and for all MNLI labels.

\begin{table}
	\small
	\centering
	\setlength\tabcolsep{5pt}
	\newcolumntype{Y}{>{\centering\arraybackslash}X}
	\begin{tabularx}{\linewidth}{lYYYY}
		\toprule
		& \multicolumn{1}{c}{\textbf{Yelp}} & \multicolumn{1}{c}{\textbf{AG's}} & \multicolumn{1}{c}{\textbf{Yahoo}} & {\textbf{MNLI}} \\
		\midrule
		supervised & $44.8$ & $82.1$ & $52.5$ & $45.6$ \\
		\textsc{Pet} & $\mathbf{60.0}$ & $\mathbf{86.3}$ & $\mathbf{66.2}$ & $\mathbf{63.9}$ \\
		\textsc{Pet} + AVS & $55.2$ & $85.0$ & $58.2$ & $52.6$ \\
		\bottomrule
	\end{tabularx}
	\caption{Results for supervised learning, \textsc{Pet} and \textsc{Pet} with AVS (\textsc{Pet} + AVS) after training on 50 examples}
	\label{automatic-verbalizers}
\end{table}

\begin{table}
	\small
	\centering
	\begin{tabularx}{\linewidth}{lX}
		\toprule
		$y$ & \textbf{Top Verbalizers} \\
		\midrule
		1 & worthless, BAD, useless, appalling \\
		2 & worse, slow, frustrating, annoying\\ 
		3 & edible, mixed, cute, tasty, Okay \\
		4 & marvelous, loved, love, divine, fab \\
		5 & golden, magical, marvelous, perfection\\ 
		\bottomrule
	\end{tabularx}
	\caption{Most probable verbalizers according to AVS for Yelp with 50 training examples}
	\label{avs-yelp}
\end{table}

\end{document}